%% file: 0_conf_main.tex
\documentclass[conference]{IEEEtran}
\IEEEoverridecommandlockouts

\usepackage{amsmath,amssymb,amsfonts}
\usepackage{multirow,multicol}
\usepackage{algorithmic}
\usepackage{subcaption}
\usepackage{adjustbox}
\usepackage{graphicx}
\usepackage{booktabs}
\usepackage{textcomp}
\usepackage{hyperref}
\usepackage{xcolor}
\usepackage{cite}
\usepackage{url}

\def\BibTeX{{\rm B\kern-.05em{\sc i\kern-.025em b}\kern-.08em
    T\kern-.1667em\lower.7ex\hbox{E}\kern-.125emX}}

\usepackage{todonotes}

\begin{document}

\title{Exploring Fine-grained Retail Product Discrimination with Zero-shot Object Classification Using Vision-Language Models\\
}

\author{
\IEEEauthorblockN{Anil Osman Tur}
\IEEEauthorblockA{
    \textit{\hspace{10mm}DISI\hspace{10mm}} \\
    \textit{ University of Trento}\\ 
    Trento, Italy \\  
} 
\and
\IEEEauthorblockN{Alessandro Conti}
\IEEEauthorblockA{
    \textit{\hspace{15mm}DISI\hspace{15mm}} \\
    \textit{University of Trento} \\
    Trento, Italy \\  
}
\and
\IEEEauthorblockN{Cigdem Beyan}
\IEEEauthorblockA{
    \textit{Dep. of Computer Science} \\
    \textit{University of Verona} \\
    Verona, Italy \\  
}
\and
\IEEEauthorblockN{Davide Boscaini}
\IEEEauthorblockA{
    \textit{\hspace{5mm}Technologies of Vision\hspace{5mm}} \\
    \textit{Fondazione Bruno Kessler}\\
    Trento, Italy \\ %
}
\and
\IEEEauthorblockN{Roberto Larcher}
\IEEEauthorblockA{
    \textit{\hspace{10mm}Spindox Labs\hspace{10mm}} \\
    \textit{Spindox SpA}\\
    Trento, Italy \\  
}
\and
\IEEEauthorblockN{Stefano Messelodi}
\IEEEauthorblockA{
    \textit{Technologies of Vision} \\
    \textit{Fondazione Bruno Kessler} \\
    Trento, Italy \\
}
\and
\IEEEauthorblockN{Fabio Poiesi}
\IEEEauthorblockA{
    \textit{\hspace{5mm}Technologies of Vision\hspace{5mm}} \\
    \textit{Fondazione Bruno Kessler} \\
    Trento, Italy \\
}
\and
\IEEEauthorblockN{Elisa Ricci}
\IEEEauthorblockA{
    \textit{\hspace{5mm}University of Trento,\hspace{5mm}}\\
    \textit{Fondazione Bruno Kessler}\\
    Trento, Italy \\  
}
}

\maketitle

\vspace{-5pt}

\begin{abstract}
In smart retail applications, the large number of products and their frequent turnover necessitate reliable zero-shot object classification methods. The zero-shot assumption is essential to avoid the need for re-training the classifier every time a new product is introduced into stock or an existing product undergoes rebranding. In this paper, we make three key contributions. Firstly, we introduce the MIMEX dataset, comprising 28 distinct product categories. Unlike existing datasets in the literature, MIMEX focuses on fine-grained product classification and includes a diverse range of retail products. Secondly, we benchmark the zero-shot object classification performance of state-of-the-art vision-language models (VLMs) on the proposed MIMEX dataset. Our experiments reveal that these models achieve unsatisfactory fine-grained classification performance, highlighting the need for specialized approaches. Lastly, we propose a novel ensemble approach that integrates embeddings from CLIP and DINOv2 with dimensionality reduction techniques to enhance classification performance. By combining these components, our ensemble approach outperforms VLMs, effectively capturing visual cues crucial for fine-grained product discrimination. Additionally, we introduce a class adaptation method that utilizes visual prototyping with limited samples in scenarios with scarce labeled data, addressing a critical need in retail environments where product variety frequently changes. To encourage further research into zero-shot object classification for smart retail applications, we will release both the MIMEX dataset and benchmark to the research community. Interested researchers can contact the authors for details on the terms and conditions of use. The code is available: \href{https://github.com/AnilOsmanTur/Zero-shot-Retail-Product-Classification}{https://github.com/AnilOsmanTur/Zero-shot-Retail-Product-Classification}.
\end{abstract}


\vspace{-5pt}
\input{1_intro}\vspace{-2pt}
\input{2_related}\vspace{-2pt}
\input{3_dataset}\vspace{-2pt}
\input{4_method}\vspace{-2pt}
\input{5_results}\vspace{-2pt}

\section{Discussions and Conclusions}\vspace{-2pt}
This study advances zero-shot object classification in smart retail by introducing the fine-grained MIMEX dataset and exploring the use of direct image embeddings for classification. Our novel contributions include the development of a highly detailed dataset with fine-grain labels and the introduction of class adaptation through visual prototyping with limited sample sizes. We demonstrate that VLMs like CLIP, BLIP, and DINOv2 can effectively perform zero-shot classification on this dataset, which presents a significant challenge due to the fine-grained nature of the product classes.
Key findings from our experiments indicate that:
(a) Enhanced textual embeddings with detailed descriptions significantly improve classification accuracy.
(b) Visual prototypes, especially when combined with k-NN classification, offer a robust alternative to textual descriptions.
(c) An ensemble model that integrates CLIP and DINOv2 visual embeddings with PCA to reduce noise in embeddings achieves further performance gains.
Practical implications for smart retail include the ability to quickly adapt to inventory changes without retraining, improved customer experience through efficient product recognition, and the potential for cost-effective, scalable classification systems.
Future research directions involve refining prompt engineering techniques, exploring the synergy of zero-shot and few-shot learning, expanding the dataset to cover a broader product spectrum, and developing real-time classification systems for complex retail environments.\vspace{-2pt}

\section*{Acknowledgment}\vspace{-1pt}
The work is partially funded by the European Union (EU) project FSE REACT-EU, MIUR PON R\&I 2014-2020 (CCI 2014IT16M2OP005). We acknowledge the support of the MUR PNRR project iNEST-Interconnected Nord-Est Innovation Ecosystem (ECS00000043) and FAIR - Future AI Research (PE00000013), both funded by NextGenerationEU. The work was carried out in the Vision and Learning joint laboratory of FBK and UNITN.

\vspace{-5pt}
\bibliographystyle{ieeetr}
\bibliography{listbib}

\end{document}

%% file: 1_intro.tex
\section{Introduction}\label{sec:introduction}
\vspace{-1mm}
The global smart retail market is projected to reach USD 227.29 billion by 2030, growing at a compound annual growth rate of 29.1\% from 2023 to 2030~\cite{size2023share}. Innovations like Amazon GO have revolutionized retail by eliminating cashiers and automating tasks, thereby enhancing customer experience, inventory management, and personalized marketing~\cite{wankhede2018just, guimaraes2023review}.

As these technological advancements reshape the retail landscape, the need for sophisticated automated systems becomes increasingly critical. This is where zero-shot object classification comes into play. It offers a powerful solution for the dynamic and diverse environment of smart retail. Zero-shot object classification, in particular, has significantly impacted computer vision, with models like CLIP~\cite{clip} and BLIP~\cite{blip} leading the way. These models excel at classifying objects without needing labeled training data by leveraging textual embeddings and visual prototypes. Additionally, vision encoders like DINOv2~\cite{dinov2} have demonstrated strong performance in extracting robust visual features, which can be utilized to enhance the visual understanding capabilities of zero-shot classification models. The combination of these approaches is especially valuable in the dynamic and diverse environment of smart retail, where the constant evolution of product ranges necessitates efficient and accurate categorization methods.

Despite the success of these state-of-the-art models~\cite{clip,blip,dinov2} on popular benchmarks, their practical application in real-world smart retail scenarios is less explored. The lack of efficient zero-shot learning models tailored specifically for smart retail contexts presents a gap.

While studies like RetailKLIP~\cite{srivastava2023retailklip}, Self-supervised Contrastive learning~\cite{nath2022self}, TemplateFree~\cite{sun2020templatefree}, and Fragmented Target Recognition~\cite{ji2023fragmented}, have begun adapting machine learning methods for retail, they often do not fully leverage the zero-shot learning capabilities of vision-language models (VLMs).

In this paper, we aim to bridge this gap by adapting the zero-shot learning capabilities of state-of-the-art models like CLIP~\cite{clip}, BLIP~\cite{blip}, and DINOv2~\cite{dinov2} to the specific needs of smart retail in a fine-grained classification setting. Our methodology integrates both textual embeddings and visual prototypes to assess their effectiveness in this specific context. We further enhance our model's capabilities by leveraging BLIP2~\cite{li2023blip2}'s advanced captioning capabilities to generate detailed product descriptions as prompts for CLIP text embeddings. To complement textual embeddings, we explore visual embeddings and prototypes for classification. By adapting our configuration for one- and few-shot transfer capabilities, we enable image-conditioned object detection without relying on text-derived query embeddings. We use the Nearest Prototype Classifier (NPC)~\cite{kuncheva1998nearest} and k-nearest neighbors (k-NN)~\cite{bishop2006pattern} algorithm to assign images to the most similar product categories based on their visual features. Moreover, we introduce an ensemble model that merges visual embeddings from CLIP and DINOv2, enhanced by dimensionality reduction techniques, to improve classification accuracy and robustness.

To assess the effectiveness of our approach in a realistic testbed, we introduce the MIMEX dataset, a collection of product images captured in a smart retail environment. Developed as part of the Micro Market Experience (MIMEX) EU project \cite{mimex} focused on advancing smart retail technologies, the MIMEX dataset provides a fine-grained testbed for evaluating our methods.
The proposed dataset is particularly challenging for zero-shot models due to its fine-grained nature and the high similarity between different product categories. Our experiments show that our ensemble model achieves a classification accuracy improvement of 15\% over the baseline models. This significant enhancement shows the potential of our approach in real-world applications, particularly in dynamic environments like smart retail.

%% file: 2_related.tex
\section{Related Work} \label{sec:related}\vspace{-1mm}

\noindent \textbf{Vision-language models (VLMs).}
VLMs like CLIP \cite{clip}, BLIP \cite{blip} and their iterations have revolutionized zero-shot learning and open-set object classification. CLIP, in particular, has demonstrated robustness across various visual understanding tasks \cite{clip,taori2020measuring,goh2021multimodal}. These models utilize dual encoders to align visual and textual data within a shared embedding space, enhanced by techniques like cross-modal attention \cite{wangsimvlm} and multi-object representation alignment \cite{zeng2022xvlm}. Subsequent research has introduced enhancements such as learning from weak supervision \cite{li2021align, singh2022flava}. The availability of large-scale datasets like LAION-5B \cite{schuhmann2022laionb} and platforms like OpenCLIP \cite{ilharco_gabriel_2021_5143773} has facilitated the widespread adoption of VLMs.

\noindent \textbf{Zero-shot learning (ZSL).}
ZSL has been a focal area in VLM research, with models like CLIP \cite{clip} effectively transferring knowledge to downstream tasks such as classification, object detection, and segmentation without the need for fine-tuning \cite{clip,gu2021open,xu2022simple}. Advanced prompt engineering and tuning strategies have significantly enhanced their generalization capabilities in zero- and few-shot scenarios \cite{clip,zhou2022learning}. Beyond these applications, CLIP has also excelled in language-guided detection and segmentation \cite{zhou2022extract,rao2022denseclip,zhong2022regionclip}, and has demonstrated promising results in fine-grained image classification, effectively recognizing subtle distinctions between highly similar objects \cite{haixin2023marinedet,wang2023open,sain2023clip}.

\noindent \textbf{Object detection methods for retail products.}
Recent advancements in deep learning have significantly impacted retail product recognition. However, challenges persist due to product diversity and the limitations of traditional object detection methods such as YOLO \cite{redmon2016you}, SSD \cite{liu2016ssd}, and Faster R-CNN \cite{girshick2015fast}. These challenges highlight the necessity for generalizable models that can efficiently handle diverse retail environments without the need for extensive retraining \cite{wei2020deep}.

Several approaches have been proposed in the literature to address these challenges. RetailKLIP \cite{srivastava2023retailklip} employs the vision encoder of a CLIP model after a finetuning to retail products, ideal for self-checkout systems and supply chain automation. This method utilizes nearest-neighbor-based classification and eliminates the need for incremental training with new products.
Complementing this, Nath \cite{nath2022self} introduces a self-supervised Siamese network capable of verifying product presence in query images without labeled data.
Their experimental analysis, however, is restricted to 310 images collected from the conveyor belt.
The TemplateFree system \cite{sun2020templatefree} introduces a product detection method using single-shelf images, eliminating the need for additional memory. It employs zero-shot learning to enhance product segmentation and detection by segmenting shelves into layers and products.
Ji et al. \cite{ji2023fragmented} investigated recognizing fragmented targets, offering solutions for detecting damaged or obscured retail products. Their method addresses the zero-shot scenario by introducing high-level attributes. However, their experimental validation focused on a limited set of five common ingredients: cucumber, potato, tomato, eggplant, and bamboo.

While prior studies have made notable advancements in detection and classification, our approach diverges by focusing on a broader range of products with high inter-class similarity, such as different brands of chocolate bars or various flavors of chips from the same brand. Our method efficiently utilizes pre-trained VLMs for zero-shot classification and achieves fine-grained classification using visual prototypes, setting it apart from previous work.

%% file: 3_dataset.tex
\section{Dataset} \label{sec:dataset}\vspace{-1mm}

The dataset employed in this study, referred to as the MIMEX dataset, which is developed as part of the Micro Market Experience (MIMEX) EU project \cite{mimex}, includes a diverse array of 28 retail products, each representing a unique class. These classes span various categories, including chocolates, snacks, beverages, condiments, and personal care items. The dataset was self-constructed for the purposes of this research, ensuring a comprehensive representation of typical retail environments.

\noindent \textbf{Data acquisition and preprocessing.} The curation of the dataset images involved a meticulous process where hand localization was employed to identify interactions between hands and retail products. This method was chosen to simulate realistic consumer interactions with products, which are frequent in retail settings. Once hands were localized within the scene, cropping was performed around these regions to focus exclusively on the product items being handled.

The training set includes 10,357 images featuring multiple individuals, while the testing set consists of 4,920 images from an individual not present in the training set. This setup ensures that visible hands in the test set belong to a single individual exclusive to the test set. This separation aids in evaluating the generalization across unseen individuals. Additionally, this distinction aims to assess the robustness of variations in product presentation and handling.

\begin{figure}[t!]
\centering
\includegraphics[width=0.7\linewidth]{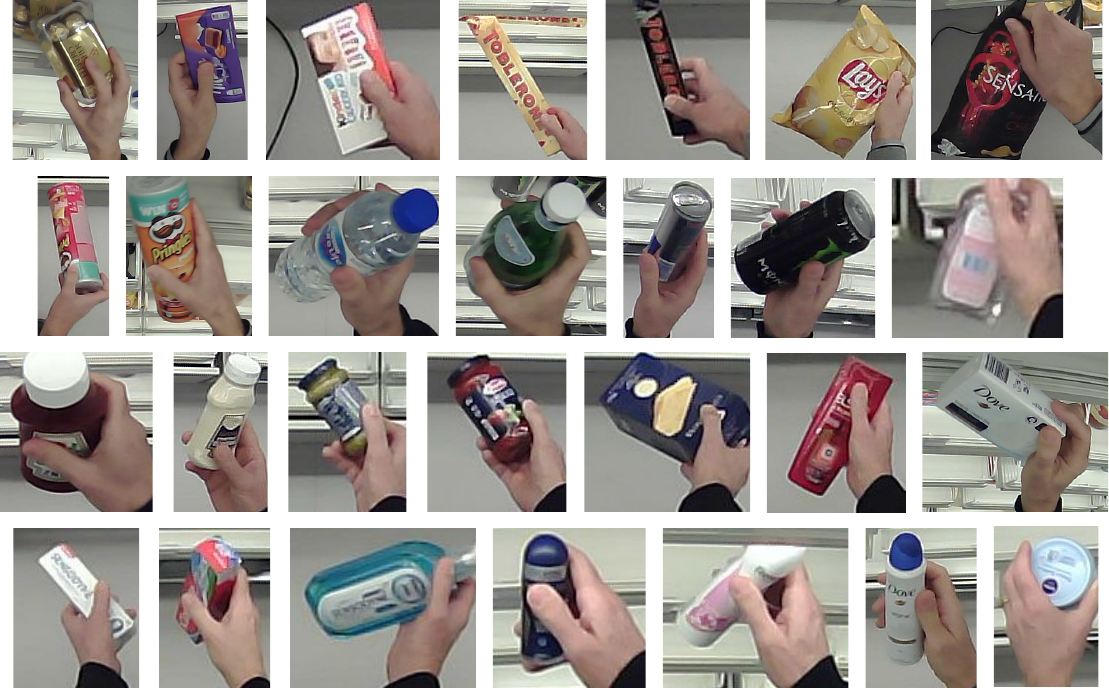}
\caption{Example images from the MIMEX dataset showcasing the cropped patches with various orientations and occlusions. The dataset includes visually similar products, such as pasta sauces, potato chips cans, and chocolates with similar packaging, highlighting the challenge of fine-grained classification in retail environments.}
\label{fig:data_ex}
\vspace{-2mm}
\end{figure}

\begin{figure}[t!]
\centering
\includegraphics[width=0.8\linewidth]{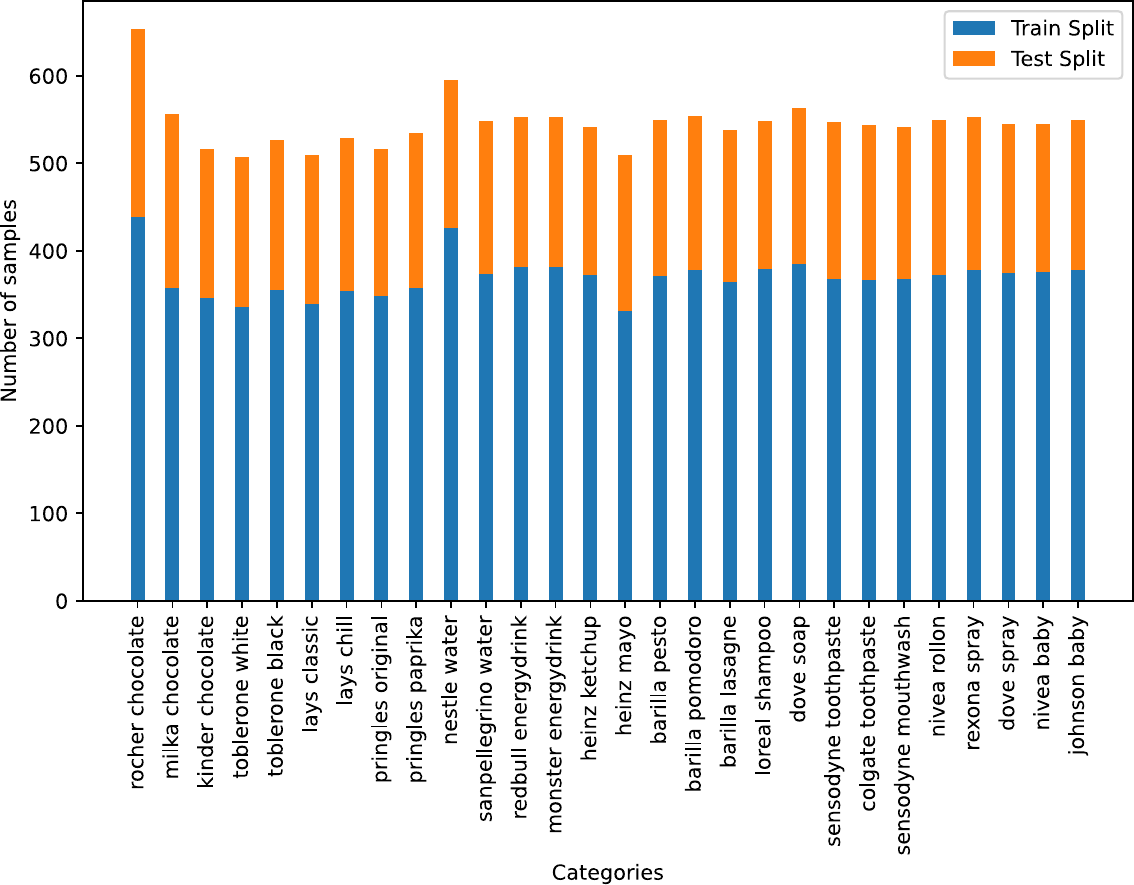}
\caption{Distribution of image samples across product categories in the MIMEX dataset, with colors indicating the allocation to the test (orange) and train (blue) splits.
}
\label{fig:data_dist}
\vspace{-5mm}
\end{figure}

\noindent \textbf{Dataset composition.} 
The categories include, but are not limited to, rocher chocolate, milka chocolate, kinder chocolate, toblerone variants, lays chips, pringles, bottled waters, energy drinks, condiments, pasta sauces, shampoos, soaps, kinds of toothpaste, mouthwashes, and deodorants. These cropped regions are visualized in Fig.~\ref{fig:data_ex}. The distribution of samples across categories and splits can be seen in Fig.~\ref{fig:data_dist}.

\noindent \textbf{Dataset challenges.} The MIMEX dataset, designed to reflect real-world retail settings, presents several challenges. Diverse lighting conditions in retail environments can significantly alter the appearance of products, complicating their recognition. Additionally, products are often partially obstructed by the client's hand or other items, which can hide important visual identifiers essential for accurate classification. The random orientation of products in consumers' hands further requires models to recognize items from multiple angles and positions. Moreover, retail environments frequently house visually similar products.

%% file: 4_method.tex
\section{Method}\label{sec:method} \vspace{-1mm}
We present an approach for retail product classification using textual embeddings and visual prototypes, leveraging self-supervised models: CLIP \cite{clip}, BLIP \cite{blip}, BLIP-2 \cite{li2023blip2}, and DINOv2 \cite{dinov2} for zero-shot classification.

\vspace{-1mm} \subsection{Preliminaries} \vspace{-1mm}
\noindent\textbf{Contrastive language-image pre-training (CLIP)} \cite{clip} has transformed open-set visual understanding with its dual-encoder framework. The image encoder converts input images ($I \in \mathbb{R}^{H \times W \times 3}$) into fixed-size patches, which, along with a learnable class token, are mapped into a unified vision-language embedding space with visual encoder $\mathbf{V}$ to produce visual features $f_I = \mathbf{V}(I) \in \mathbb{R}^d$.
Concurrently, the text encoder $\mathbf{T}$ transforms sentences into word embeddings, integrating a class token to generate a feature matrix. This matrix is processed to derive textual features $f_{t} = \mathbf{T}(\mathcal{S})$. CLIP is trained with a contrastive loss to enhance the similarity of matching text-image pairs and reduce the similarity of non-matching ones. For classification, it uses textual prompts to generate specific text features and compute the prediction by calculating the distance to an image feature as:
\begin{equation}
\label{eq:primary}
\mathcal{P}(y|I) = \frac{\mathrm{exp}(\texttt{sim}(f_I, f_t^y)/\tau)}{\sum_{i=1}^{K} \mathrm{exp}(\texttt{sim}(f_I, f_t^i)/\tau)}.
\end{equation}

\noindent\textbf{Bootstrapped language image pre-training (BLIP)} \cite{blip} extends VLM capabilities to image captioning tasks, utilizing a Vision Transformer (ViT) backbone pre-trained on the COCO dataset \cite{coco}. This architecture enhances the descriptive power of generated captions, facilitating the creation of accurate textual embeddings. Building upon BLIP \cite{blip}, BLIP-2 \cite{li2023blip2} features a CLIP-like image encoder, a querying transformer (Q-Former), integrates a large language model (LLM) with 2.7 billion parameters.
The Q-Former acts as a bridge, mapping the query tokens to the query embeddings, enabling the model to predict the next text token and generate contextually relevant captions or descriptions.

\noindent\textbf{Self-DIstillation with NO labels (DINOv2)} \cite{dinov2} utilizes a ViT architecture trained self-supervised on a vast image collection, yielding robust visual features for fine-grained visual understanding \cite{he2023improved}. It processes images as sequences of fixed-size patches, incorporating a [CLS] token for classification tasks.

\vspace{-1mm} \subsection{Our approach} \vspace{-1mm}
\noindent\textbf{Textual embedding enhancement.}
Our method enhances zero-shot classification by generating descriptive prompts that capture the essence of product classes. We generate a set of textual prototypes by merging class names and templates (\textit{e.g.}, ``A photo of a [$c$],'' where $c$ is the class name) to enhance the model's capabilities to classify unknown objects in images.
We employ an ensemble of prompt templates, such as ``A cropped photo of a [$c$]'' and ``A centered photo of a [$c$] consumer product'', to generate more precise textual descriptors, improving classification by the CLIP model. 
Additionally, we used BLIP2 to generate captions from images, which provided detailed descriptions of products. The generated captions were used as prompts for the CLIP.

\begin{figure*}[t!]
\centering
\includegraphics[width=0.65\linewidth]{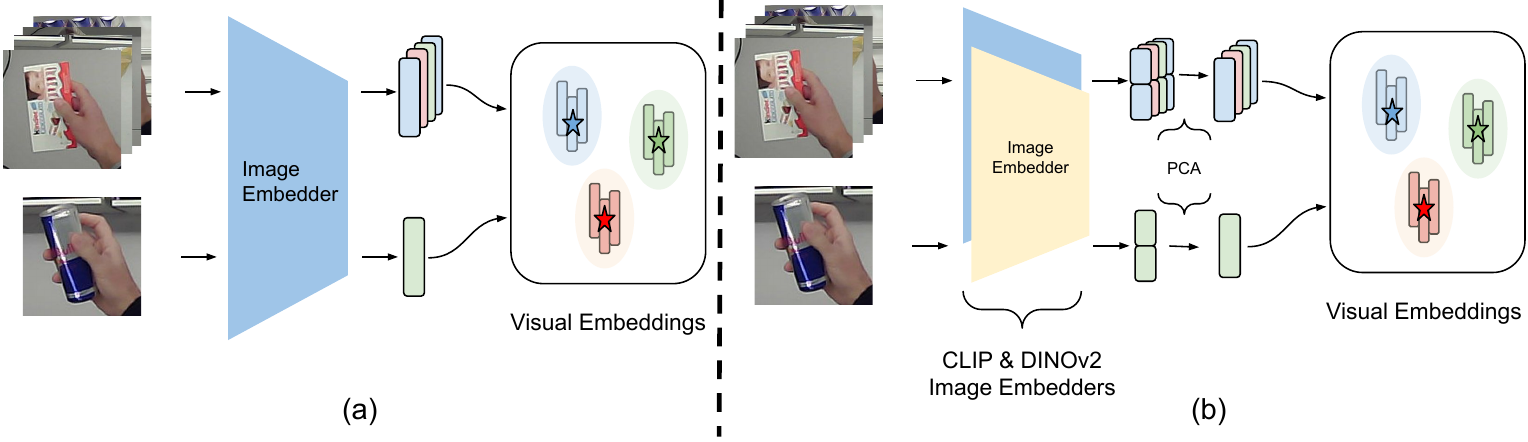}
\caption{
(a) Prototype generation: This panel illustrates the process of creating prototypes by extracting and refining visual embeddings from image data.
(b) Integration of CLIP and DINOv2 features to create visual prototypes: This panel shows how visual embeddings from CLIP and DINOv2 are combined and the use of PCA to reduce dimensionality and enhance the clarity of the prototypes, facilitating improved understanding within contexts.
}
\label{fig:embeddings}
\vspace{-5mm}
\end{figure*}

\noindent\textbf{Classification with visual embeddings and prototypes.}
To enhance our textual embeddings, we incorporate one or few-shot capabilities using direct image-based embeddings for classification (Fig. \ref{fig:embeddings}). This allows for image-conditioned one-shot object detection, ideal for objects that are challenging to describe textually. We utilize the Nearest Prototype Classifier (NPC) \cite{kuncheva1998nearest}, which assigns an image to the class of its nearest prototype—defined as the mean of the class's image representations. This method assumes hyperspherical class distributions of equal volume and prior probabilities.

Additionally, we adapt the k-nearest neighbors (k-NN) algorithm to the high-dimensional feature space from CLIP and DINOv2 models. By calculating the Euclidean distance between an unlabeled image's feature vector and those of labeled images, we determine the class based on the most frequent label among the k closest labeled images, optimizing k for best classification accuracy.

\noindent\textbf{Utilization of CLIP and DINOv2 for image embeddings.}
Integrating CLIP and Dinov2 plays a crucial role in our approach, as shown in Fig.~\ref{fig:embeddings}(a). CLIP encodes visual information into embeddings by leveraging a vast corpus of image-text pairs, enabling generalization from textual to visual representations. After generating visual embeddings with CLIP and DINOv2, we apply Principal Component Analysis (PCA) to reduce the dimensionality and noise in the embeddings \cite{rosipal2001kernel}. This step is vital for improving the model's efficiency and accuracy, especially when processing large-scale image datasets for precise classification.

\noindent\textbf{Implementation details.}
Our approach utilizes the ViT-B/32@224 variant of the CLIP backbone, which is pre-trained on the LAION-2B subset of the LAION-5B dataset \cite{schuhmann2022laionb}. We also use the BLIP base model, pre-trained on COCO \cite{coco}, and the BLIP-2 model, which is integrated with OPT-2.7b \cite{zhang2022opt} to enhance its capabilities. Additionally, we employ the baseline ViT version of DINOv2 \cite{imagenet22} for extracting visual features.

%% file: 5_results.tex
\section{Experiments}\label{sec:experiments}\vspace{-2mm}
Our experiments are designed to evaluate the zero-shot inference capabilities of our model for open-set object classification. Additionally, we assess the model's performance in few-shot object classification and conduct ablation studies of our approach.

\noindent \textbf{Evaluation protocol and metrics.}
Our experimental evaluation focused on the following variables: the number of textual prompts, the number of neighbors (k) in the k-NN algorithm for visual prototype generation, prototype sample size, and PCA dimension to reduce. We measured Top-1 accuracy to assess classification performance across different setups. We employed 2-fold cross-validation, alternating between test and train splits to evaluate the dataset and model robustness thoroughly. After visually inspecting the dataset, we discarded hard samples such as those totally occluded by hand, other objects, or saturated because of light glare, and created a cleaner version, referred to as \textit{Cleaned} in the tables. The original version is called \textit{Base}. This cleaning process aimed to improve the quality and consistency of the data used in our experiments.

\noindent \textbf{Classification with textual embeddings.}
Our zero-shot classification approach leverages textual embeddings to enhance accuracy in real-world retail settings. We discovered that BLIP-generated captions significantly improve classification, highlighting the importance of precise and detailed class labels. Our methodology evolved from a basic CLIP template, referred to as $baseline$, to an expanded set of 44 templates, termed $multiple$ prompts. Through analysis, we identified the most effective prompts, designated as $selected$ templates. Table \ref{tab:clip_zeroshot} compares the performance of these templates across Test, Train, and All Data splits.
\input{tables/clip_zeroshot} 

Table \ref{tab:blip_captions} presents the results of our experiments using BLIP-generated captions as textual representations. We generated captions separately for the train and test splits and averaged the caption embeddings to form text descriptors. The train split descriptions were more detailed than those from the test split, likely due to the greater diversity of individuals and product interactions in the training set. Further testing with BLIP2 showed that its captions better described the images and improved classification scores. These findings emphasize the importance of well-designed textual prompts and the potential of advanced captioning models like BLIP for enhancing zero-shot classification performance in smart retail environments. However, the challenge of describing specialized product packaging led us to explore few-shot scenarios.
\input{tables/blip_captions}

\noindent \textbf{Classification with visual prototypes.}
Our experiments with visual prototypes involve creating prototypes by averaging sample values and applying the k-NN algorithm with various numbers of neighbors ($k$) from 1 to 11. This approach helps us understand the data distribution and the descriptive power of CLIP's visual feature space. Table \ref{tab:clip_embeddings} demonstrates the effectiveness of visual prototypes for classification, with CLIP achieving high accuracy scores by capturing distinct visual features that differentiate between classes.
\input{tables/clip_embeddings}

We explore prototype selection with different sample sizes using the cleaned dataset (Table \ref{tab:clip_prototypes}). Our method maintains a high accuracy of 89.90\% with just 50 samples per class, only a 4\% decrease compared to the original setting. This shows the adaptability of our model to novel product categories with minimal supervision.
\input{tables/clip_prototypes}

To further enhance classification performance, we combine visual features from CLIP and DINOv2 models (Table \ref{tab:clip_dino_embeddings}). This ensemble approach leverages the complementary strengths of both models, surpassing their individual accuracies. Additionally, applying PCA reduces noise and increases the descriptive ability of the features.

\input{tables/clip_dino_embeddings}

\noindent \textbf{Observations and insights.}
Our findings reveal that textual descriptions provide superior open-set classification capabilities compared to visual prototypes. However, optimal classification performance hinges on the careful selection of descriptors and the use of an effective feature extractor. These results show the potential of advanced VLMs in addressing the unique challenges of smart retail product classification.

\noindent \textbf{Visualizations of feature Space.}
To better understand the feature space and evaluate our model's performance, we employ Uniform Manifold Approximation and Projection (UMAP) \cite{umap} for visualizing the high-dimensional embeddings from the CLIP \cite{clip} model. UMAP effectively preserves the overall data structure while revealing local patterns and clusters.
\begin{figure}[!t]
\centering
\includegraphics[width=0.9\linewidth]{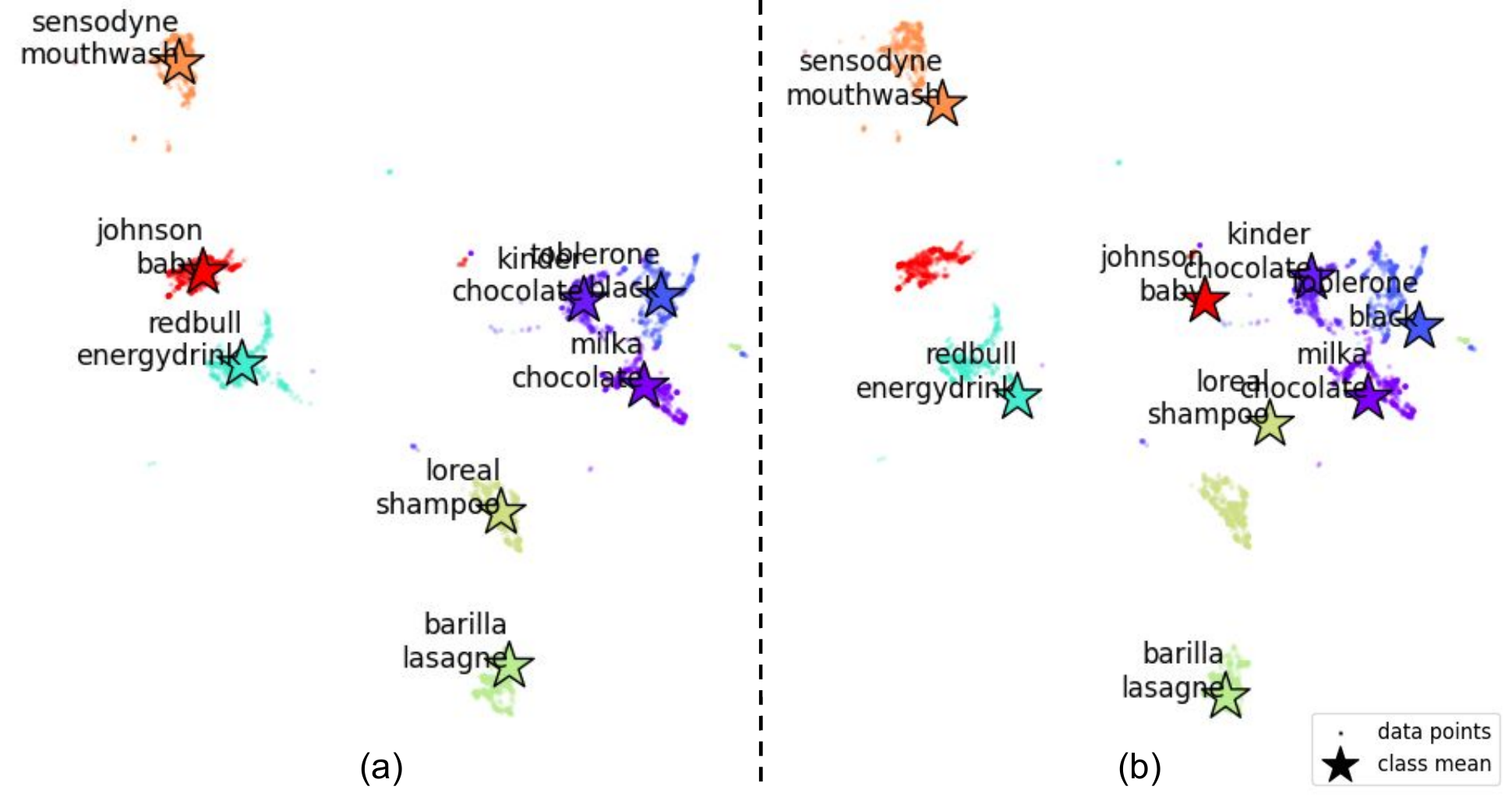}
\caption{(a) UMAP visualization of CLIP's feature space on the MIMEX dataset, with distinct class representations in different colors and class centroids marked by stars. (b) Visualization of class centroids based on prompt predictions, emphasizing the central points of each class.}
\vspace{-5mm}
\label{fig:umap_clip}
\end{figure}

Figure \ref{fig:umap_clip} illustrates the UMAP visualization of eight randomly selected classes from the MIMEX dataset, highlighting how the zero-shot CLIP model effectively handles class separation based on textual or visual cues.

%% file: tables/clip_zeroshot.tex
\begin{table}[t]
    \centering
    \tabcolsep 3pt
    \caption{Accuracy (\%) using textual prompts for zero-shot classification with CLIP.}
    \label{tab:clip_zeroshot}
    \begin{adjustbox}{width=0.5\columnwidth,center}
    \begin{tabular}{llccc}
        \toprule
        ~ & Prompts & Test   & Train  & All Data \\
        \toprule
        \multirow{3}{*}{Base} & baseline & 40.22 & 46.44 & 44.42 \\
        ~ & multiple              & 41.17 & 47.30 & 45.31 \\
        ~ & selected              & \textbf{42.29} & \textbf{48.65} & \textbf{46.59} \\
        \toprule
        \multirow{3}{*}{Cleaned} & baseline & 48.96  & 44.29  & 47.51 \\
        ~ & multiple                 & 49.65  & 45.12  & 48.24 \\
        ~ & selected                 & \textbf{50.33}  & \textbf{45.97}  & \textbf{48.98} \\
        \bottomrule
    \end{tabular}
    \end{adjustbox}
    \vspace{-6mm}
\end{table}

%% file: tables/blip_captions.tex
\begin{table}[t]
    \centering
    \caption{Accuracy (\%) using generated BLIP captions as textual prompts for zero-shot classification with CLIP.}
    \label{tab:blip_captions}
    \begin{adjustbox}{width=0.8\columnwidth,center}
    \begin{tabular}{lllccc}
        \toprule
        ~ & ~ & Captions & Test   & Train  & All Data \\
        \toprule
        \multirow{6}{*}{Base} & \multirow{3}{*}{BLIP}  & train  & 54.35 & 48.42 & 52.44 \\
        ~                     &                     ~  & test   & 49.41 & 55.16 & 51.26 \\
        ~                     &                     ~  & all    & 54.35 & 48.42 & 52.44 \\ \cline{2-6}
        ~                     & \multirow{3}{*}{BLIP2} & train  & 57.69 & 50.02 & 55.22 \\
        ~                     & ~                      & test   & \textbf{60.45} & \textbf{56.42} & \textbf{59.15} \\
        ~                     & ~                      & all    & 57.69 & 50.02 & 55.22 \\
        \toprule
        \multirow{6}{*}{Cleaned} & \multirow{3}{*}{BLIP}  & train  & 55.92 & 52.72 & 54.93 \\
        ~                        & ~                      & test   & 53.24 & 58.96 & 55.01 \\
        ~                        & ~                      & all    & 55.92 & 52.72 & 54.93 \\ \cline{2-6} 
        ~                        & \multirow{3}{*}{BLIP2} & train  & 58.47 & 53.15 & 56.82 \\
        ~                        & ~                      & test   & \textbf{62.30} & \textbf{60.43} & \textbf{61.72} \\
        ~                        & ~                      & all    & 58.47 & 53.15 & 56.82 \\
        \bottomrule
    \end{tabular}
    \end{adjustbox}
    \vspace{-3mm} \vspace{2pt}
\end{table}

%% file: tables/clip_embeddings.tex
\begin{table}[t]
    \centering
    \caption{Accuracy (\%) achieved by employing CLIP visual embeddings. \textit{means} refers to the average embedding vectors used as class prototypes.}
    \label{tab:clip_embeddings}
    \begin{adjustbox}{width=0.8\columnwidth,center}
    \begin{tabular}{llccc}
        \toprule
        ~                     & ~            & Train $\rightarrow$ Test & Test $\rightarrow$ Train & Average \\
        \toprule
\multirow{6}{*}{Base}    & means & 82.91 & 87.60 & 85.26 $\pm$ 3.3 \\
                         &  k=1      & 85.43 & 87.87 & 86.65 $\pm$ 1.7 \\
                         &  k=3      & 86.52 & 89.28 & 87.90 $\pm$ 2.0 \\
                         &  k=5      & 87.15 & 89.46 & 88.31 $\pm$ 1.6 \\
                         &  k=7      & 87.44 & 89.68 & 88.56 $\pm$ 1.6 \\
                         &  k=11     & \textbf{88.13} & \textbf{89.90} & \textbf{89.02 $\pm$ 1.3} \\ 
                         \toprule
\multirow{6}{*}{Cleaned} & means & 90.49 & 92.98 & 91.74 $\pm$ 1.8 \\
                         &  k=1      & 91.44 & 92.11 & 91.78 $\pm$ 0.5 \\
                         &  k=3      & 92.82 & 93.22 & 93.02 $\pm$ 0.3 \\
                         &  k=5      & 93.46 & 93.38 & 93.42 $\pm$ 0.1 \\
                         &  k=7      & 93.78 & 93.52 & 93.65 $\pm$ 0.2 \\
                         &  k=11     & \textbf{94.06} & \textbf{93.82} & \textbf{93.94 $\pm$ 0.2} \\
        \bottomrule
    \end{tabular}
    \end{adjustbox}
    \vspace{-6mm}
\end{table}

%% file: tables/clip_prototypes.tex
\begin{table}[t]
    \centering
    \caption{Accuracy (\%) using CLIP visual embeddings with prototypes. \textit{Sample size} refers to the number of data used to create each prototype.}
    \label{tab:clip_prototypes}
    \begin{adjustbox}{width=0.85\columnwidth,center}
    \begin{tabular}{cccc}
        \toprule
        Sample Size & Train $\rightarrow$ Test & Test $\rightarrow$ Train & Average \\
        \toprule
        50 & 89.53 & 90.27 & 89.90 $\pm$ 0.5 \\
        25 & 88.22 & 88.29 & 88.26 $\pm$ 0.0 \\
        20 & 87.35 & 87.28 & 87.32 $\pm$ 0.0 \\
        15 & 86.42 & 86.25 & 86.34 $\pm$ 0.1 \\
        10 & 84.66 & 83.47 & 84.07 $\pm$ 0.8 \\
       \bottomrule
    \end{tabular}
    \end{adjustbox}
    \vspace{-2mm}
\end{table}

%% file: tables/clip_dino_embeddings.tex
\begin{table}[t]
    \centering
    \tabcolsep 3pt
    \caption{Accuracy (\%) using visual embeddings and ensemble model.}
    \label{tab:clip_dino_embeddings}
    \begin{adjustbox}{width=0.9\columnwidth,center}
    \begin{tabular}{llccc}
        \toprule
        & Method & Train $\rightarrow$ Test & Test $\rightarrow$ Train & Average\\
        \toprule
        \multirow{5}{*}{Base} & DINOv2                    & 79.59          & 80.26          & 79.93 $\pm$ 0.5 \\
        ~                     & CLIP                      & 85.51          & \textbf{89.54} & 87.53 $\pm$ 2.8 \\
        ~                     & CLIP + DINOv2             & 86.22          & 89.28          & 87.75 $\pm$ 2.2 \\
        ~                     & CLIP + DINOv2 + PCA(1024) & 86.22          & 89.28          & 87.75 $\pm$ 2.2 \\
        ~                     & CLIP + DINOv2 + PCA(512)  & \textbf{86.26} & 89.26 & \textbf{87.76 $\pm$ 2.1} \\
        \toprule
        \multirow{5}{*}{Cleaned} & DINOv2                    & 83.70          & 84.38          & 84.04 $\pm$ 0.5 \\
        ~                        & CLIP                      & 90.49          & 92.98          & 91.74 $\pm$ 1.8 \\
        ~                        & CLIP + DINOv2             & \textbf{90.86} & \textbf{93.60} & \textbf{92.23 $\pm$ 1.9} \\
        ~                        & CLIP + DINOv2 + PCA(1024) & 90.81          & 93.12          & 91.97 $\pm$ 1.6 \\
        ~                        & CLIP + DINOv2 + PCA(512)  & 90.79          & 93.12          & 91.96 $\pm$ 1.6 \\
        \bottomrule
    \end{tabular}
    \end{adjustbox}
    \vspace{-2mm}
\end{table}